\documentclass{article}

\usepackage{amsmath,amsfonts,bm}
\usepackage{microtype}
\usepackage{graphicx}
\usepackage{subfigure}
\usepackage{booktabs} \usepackage[dvipsnames]{xcolor}
\usepackage{xspace}
\usepackage{enumitem}

\usepackage{hyperref}

\newcommand\sA{\ensuremath{\mathcal{A}}}

\newcommand\sD{\ensuremath{\mathcal{D}}}

\newcommand\sG{\ensuremath{\mathcal{G}}}

\newcommand\sW{\ensuremath{\mathcal{W}}}
\newcommand\sX{\ensuremath{\mathcal{X}}}
\newcommand\sY{\ensuremath{\mathcal{Y}}}

\newcommand\R{\ensuremath{\mathbb{R}}}        \newcommand{\bone}{\mathbf{1}}  \newcommand\refeqn[1]{(\ref{eqn:#1})}

\newcommand\refsec[1]{Section~\ref{sec:#1}}

\newcommand\reffig[1]{Figure~\ref{fig:#1}}

\newcommand\reftab[1]{Table~\ref{tab:#1}}
\newcommand\refapp[1]{Appendix~\ref{sec:#1}}

\newcommand\refalg[1]{Algorithm~\ref{alg:#1}}

      \newcommand{\E}{\mathbb{E}}  %

\newcommand{\ours}{\textsc{Jtt}\xspace}

\newcommand{\spuattribute}{a}

\newcommand{\pup}{\hat{f}_{\text{id}}}
\newcommand{\pdown}{\hat{f}_{\text{final}}}

\newcommand{\errors}{E}
\newcommand{\spuoracle}{\textsc{Upsample minority}\xspace}
\newcommand{\upweightfactor}{\lambda_\text{up}}

\usepackage[accepted]{icml2021}

\icmltitlerunning{Just Train Twice: Improving Group Robustness without Training Group Information}

\begin{document}

\twocolumn[

\icmltitle{Just Train Twice: Improving Group Robustness \\ without Training Group Information}

\icmlsetsymbol{equal}{*}

\begin{icmlauthorlist}
\icmlauthor{Evan Zheran Liu}{equal,stan}
\icmlauthor{Behzad Haghgoo}{equal,stan}
\icmlauthor{Annie S. Chen}{equal,stan}
\icmlauthor{Aditi Raghunathan}{stan}
\icmlauthor{Pang Wei Koh}{stan}
\icmlauthor{Shiori Sagawa}{stan}
\icmlauthor{Percy Liang}{stan}
\icmlauthor{Chelsea Finn}{stan}
\end{icmlauthorlist}

\icmlaffiliation{stan}{Stanford University}

\icmlcorrespondingauthor{Evan Zheran Liu}{evanliu@cs.stanford.edu}

\icmlkeywords{Machine Learning, ICML}

\vskip 0.3in
]

\printAffiliationsAndNotice{\icmlEqualContribution} 

\begin{abstract}
Standard training via empirical risk minimization (ERM) can produce models that achieve high accuracy on average but low accuracy on certain groups, especially in the presence of spurious correlations between the input and label.
Prior approaches that achieve high worst-group accuracy, like group distributionally robust optimization (group DRO) require expensive group annotations for each training point, whereas approaches that do not use such group annotations typically achieve unsatisfactory worst-group accuracy.
In this paper, we propose a simple two-stage approach, \ours, that first trains a standard ERM model for several epochs, and then trains a second model that upweights the training examples that the first model misclassified. Intuitively, this upweights examples from groups on which standard ERM models perform poorly, leading to improved worst-group performance. Averaged over four image classification and natural language processing tasks with spurious correlations, \ours closes 75\% of the gap in worst-group accuracy between standard ERM and group DRO, while only requiring group annotations on a small validation set in order to tune hyperparameters. 
\end{abstract} \section{Introduction}

The standard approach of empirical risk minimization (ERM)---training machine learning models to minimize average training loss---can produce models that achieve low test error on average but still incur high error on certain groups of examples~\citep{hovy2015, blodgett2016, tatman2017, hashimoto2018repeated, duchi2019distributionally}.
These performance disparities across groups can be especially pronounced in the presence of \emph{spurious correlations}.
For example, in the task of classifying whether an online comment is toxic, the training data is often biased so that mentions of particular demographics (e.g., certain races or religions) are correlated with toxicity.
Models trained via ERM then associate these demographics with toxicity and thus perform poorly on groups of examples in which the correlation does not hold, such as non-toxic comments mentioning a particular demographic \citep{borkan2019nuanced}.
Similar performance disparities due to spurious correlations occur in many other applications, including other language tasks, facial recognition, and medical imaging~\cite{gururangan2018annotation, mccoy2019right, badgeley2019deep, sagawa2020group, oakden2020hidden}.

Following prior work, we formalize this setting by considering a set of pre-defined groups (e.g., corresponding to different demographics) and seeking models that have low worst-group error \citep{sagawa2020group}.
Previous approaches typically require annotations of the group membership of each training example \citep{sagawa2020group, goel2020model, zhang2020coping}. 
While these approaches have been successful at improving worst-group performance, the required training group annotations are often expensive to obtain; for example, in the toxicity classification task mentioned above, each comment has to be annotated with all the demographic identities that are mentioned.

In this paper, we propose a simple algorithm, \ours (Just Train Twice), for improving the worst-group error \emph{without training group annotations}, instead only requiring group annotations on a much smaller validation set to tune hyperparameters.
\ours is composed of two stages: we first identify training examples that are misclassified by a standard ERM model, and then we train the final model by upweighting the examples identified in the first stage. 
Intuitively, this procedure exploits the observation that sufficiently-regularized ERM models tend to incur high worst-group training error (and subsequently high worst-group test error).
This makes selecting misclassified examples an effective heuristic for identifying examples from groups that ERM models fail on, such as minority groups. Since the final classifier upweights such examples, it performs better on such groups and achieves better minority group performance.

We evaluate \ours on two image classification datasets with spurious correlations, Waterbirds \citep{wah2011cub,sagawa2020group} and CelebA \citep{liu2015deep} and two natural language processing datasets, MultiNLI \citep{williams2018broad} and CivilComments-WILDS \citep{borkan2019nuanced,koh2021wilds}.
We use the versions of Waterbirds, CelebA, and MultiNLI from \citet{sagawa2020group}, where in Waterbirds, the label \emph{waterbird} or \emph{landbird} spuriously correlates with water in the background; in CelebA, the label \emph{blond} or \emph{non-blond} spuriously correlates with binary gender; and in MultiNLI, the label spuriously correlates with the presence of negation words.
In CivilComments-WILDS, where the input is online comments, the label \emph{toxic, non-toxic} spuriously correlates with the mention of particular demographics, as discussed above.
Our method outperforms ERM on all four datasets, with an average worst-group accuracy improvement of 16.2\%, while maintaining competitive average accuracy (only 4.2\% worse on average).
Furthermore, despite having no group annotations during training, \ours closes 75\% of the gap between ERM and group DRO, which uses complete group information on the training data.

We then empirically analyze \ours.
First, we analyze the examples identified by \ours and show that \ours upweights groups on which standard ERM models perform poorly, e.g., minority groups that do not have the spurious correlation (such as waterbirds on land in the Waterbirds dataset).
Second, we show that having validation group annotations is essential for hyperparameter tuning for \ours and other related algorithms.

Finally, we compare \ours with the distributionally robust optimization (DRO) algorithm that minimizes the conditional value at risk (CVaR).
CVaR DRO aims to train models that are robust to a wide range of potential distribution shifts by minimizing the worst-case loss over all subsets of the training set of a certain size~\citep{duchi2019distributionally}.
This objective does not require training group annotations, and it can be optimized by dynamically upweighting training examples with the highest losses in each minibatch \citep{levy2020large}.
Though CVaR DRO and \ours share conceptual similarities---they both upweight high loss training points and do not require training group information---the difference is that \ours upweights a static set of examples, while CVaR DRO dynamically re-computes which examples to update.
Empirically, we find that \ours empirically substantially outperforms CVaR DRO on worst-group accuracy in the datasets we tested. \section{Related Work}
In this paper, we focus on group robustness (i.e., training models that obtain good performance on each of a set of predefined groups in the dataset), though other notions of robustness are also studied, such as adversarial examples~\citep{biggio2013evasion,szegedy2014intriguing} or domain generalization~\citep{blanchard2011generalizing, muandet2013domain}.
Approaches for group robustness fall into the two main categories we discuss below.

\paragraph{Robustness using group information.}
Several approaches leverage group information during training, either to combat spurious correlations or handle shifts in group proportions between train and test distributions. 
For example, \citet{mohri2019agnostic, sagawa2020group, zhang2020coping} minimize the worst-group loss during training; \citet{goel2020model} synthetically expand the minority groups via generative modeling; \citet{shimodaira2000improving, byrd2019effect, sagawa2020overparameterization} reweight or subsample to artificially balance the majority and minority groups; \citet{cao2019learning, cao2020heteroskedastic} impose heavy Lipschitz regularization around minority points.
These approaches substantially reduce worst-group error, but obtaining group annotations for the entire training set can be extremely expensive.

Another line of work studies worst-group performance in the context of fairness~\citep{hardt2016, woodworth2017, pleiss2017, agarwal2018reductions, khani2019mwld}.
While these works also aim to improve the worst-group loss, they explicitly focus on equalizing the loss across all groups.

\paragraph{Robustness without group information.}
We focus on the setting where group annotations are expensive and unavailable on the training data, and potentially only available on a much smaller validation set.
Many approaches for this setting fall under the general DRO framework, where models are trained to minimize the worst-case loss across all distributions in a ball around the empirical distribution~\citep{bental2013robust, lam2015quantifying, duchi2016, namkoong2017variance, oren2019drolm}.~\citet{pezeshki2020gradient} modify the dynamics of stochastic gradient descent to avoid learning spurious correlations.~\citet{sohoni2020no} automatically identify groups based by clustering the data points.~\citet{kim2019multiaccuracy} propose an auditing scheme that searches for high-loss groups defined by a function within a pre-specified complexity class and postprocess the model to minimize discrepancies identified by the auditor.~\citet{khani2019mwld} minimize the variance in the loss across all data points to encourage lower discrepancy in the losses across all possible groups. 
Another approach is to directly learn how to reweight the training examples either using small amounts of metadata~\citep{shu2019meta} or automatically via meta-learning~\citep{ren2018reweighting}.

Most closely related to \ours are several approaches that also train a pair of models, where the performance of the first model is used to help train the second model \citep{yaghoobzadeh2019increasing, utama2020towards, nam2020learning}.
We compare \ours with one such approach, called Learning from Failure (LfF) \citep{nam2020learning}.
In LfF, the first model is intentionally biased and tries to identify minority examples where the spurious correlation does not hold.
The identified examples are then upweighted while training the second model.
This approach interleaves the updates of both models and requires an intentional biasing with the the first model.
In contrast, our approach of \ours is simpler, though conceptually similar: we only identify points to upweight once (i.e. no interleaved updates which generally destabilize training), and we just perfom standard ERM with regularization to identify points without any artificial biasing.
Empirically, despite its simplicity, \ours performs better than LfF.

Concurrently, \citet{creager2021environment} also proposed a method that leverages similar intuition to \ours.
This work first uses the errors of a standard ERM model to infer group labels, similar to \ours.
Then, they learn a model that is invariant to the predicted labels. \section{Preliminaries}\label{sec:preliminaries}

\subsection{Problem Setup}
We consider the setting of classifying an input $x \in \sX$ as a label $y \in \sY$. We are given $n$ training points $\{(x_1, y_1), \hdots, (x_n, y_n) \}$. Our goal is to learn a model $f_\theta: \sX \rightarrow \sY$, parameterized by $\theta \in \Theta$. We measure performance across a set of pre-defined groups $\sG$. Each point $(x, y)$ belongs to some group $g \in \sG$ and we evaluate classifiers on their worst-group error defined as follows: 
\begin{align}
\label{eq:wgerror}
    \max_{g \in \sG} \E \left[\ell_{0-1}(x, y; \theta) \mid g\right],
\end{align}
where $l_{0-1}(x, y ; \theta) = \bone[f_\theta(x) \neq y]$ is the 0-1 loss.

We are interested in the setting where we do not have group annotations on training points because they are expensive to obtain. Our goal is to achieve good worst-group error at test time without training group annotations. However, we are given a small validation set of $m$ points with group annotations $\{ (x_1, y_1, g_1), \hdots, (x_m, y_m, g_m)\}$. These group annotations allow us to compute the worst-group validation error, which we use to tune hyperparameters.

\begin{figure}
    \centering
    \vspace{-0.1cm}
    \includegraphics[width=0.9\columnwidth]{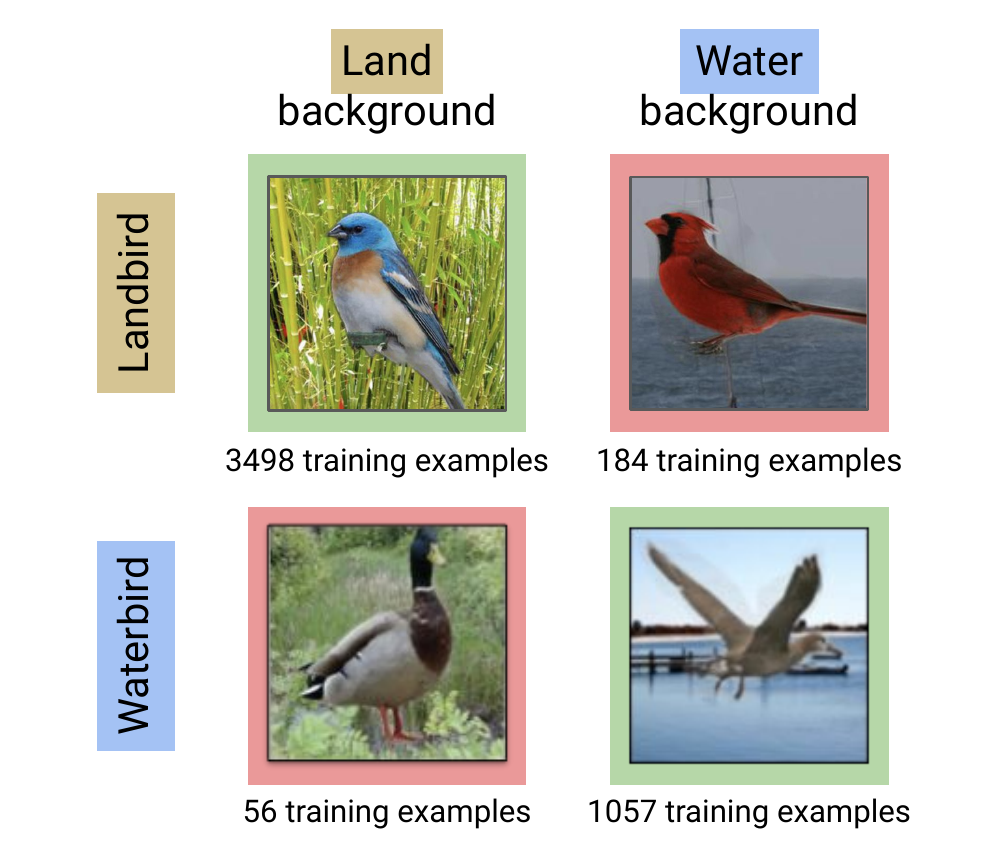}
    \vspace{-0.3cm}
    \caption{
        The four groups on the Waterbirds dataset, formed by the background spurious attribute and bird type label.
        Most training examples belong to the groups where the background matches the bird type (highlighted in green), while only a small fraction belong to the groups where the background does not match the bird type (highlighted in red).
    }
    \label{fig:waterbird_groups}
\end{figure}

\paragraph{Groups based on spurious correlations.}
In our experiments, we primarily consider the setting where each group $g = (\spuattribute, y) \in \sG$ is defined by the label $y$ and a spurious attribute $a \in \sA$ that spuriously correlates with the label (i.e., $\sG = \sA \times \sY)$. \reffig{waterbird_groups} illustrates the four groups on the Waterbirds dataset, where the background spuriously correlates with the label.

\subsection{Comparisons}
Here, we describe four other algorithms that we use as comparisons in this paper:
(i) Empirical risk minimization (ERM), which is the standard approach for training machine learning models by minimizing the average training loss (\refsec{erm}).
(ii) A distributionally robust optimization (DRO) method for minimizing the conditional value at risk (CVaR), which seeks to minimize error over all groups above a certain size \citep{duchi2019distributionally}, and is a natural approach to training models with low worst-group error without group annotations (\refsec{cvar_dro_prelim}).
(iii) Learning from Failure (LfF) \citep{nam2020learning}, a recent approach that is conceptually similar to \ours (\refsec{lff}).
(iv) Group DRO \citep{sagawa2020group}, 
which---unlike all of the preceding methods---uses training group annotations, and can therefore be considered as an oracle method that upper bounds the performance we might expect from methods that do not use training group annotations (\refsec{group_dro}).

\subsubsection{Empirical risk minimization (ERM)}\label{sec:erm}
 Empirical risk minimization minimizes the average training loss across training points. Given a loss function $\ell(x, y; \theta) : \sX \times \sY \times \Theta \to \R_+$ (e.g. cross-entropy loss), ERM minimizes the following objective: 
\begin{align}\label{eqn:erm}
    J_{\text{ERM}}(\theta) = \frac{1}{n} \sum \limits_{i=1}^n \ell(x_i, y_i; \theta).
\end{align}

\subsubsection{Distributionally robust optimization of the conditional value at risk (CVaR DRO)}\label{sec:cvar_dro_prelim}
Instead of minimizing the expected loss over the empirical training distribution, distributionally robust learning algorithms define an uncertainty set over distributions that are within some distance of the empirical training distribution, and then minimize the expected loss over the worst-case distribution in this uncertainty set \citep{duchi2019distributionally}.

In this paper, we study a classic instance of this type of worst-case loss known as the conditional value at risk (CVaR) at level $\alpha \in (0,1]$, 
which corresponds to an uncertainty set that contains all $\alpha$-sized subpopulations of the training distribution \citep{rockafellar2000optimization}.
The idea is that the worst loss over $\alpha$-sized subpopulations upper bounds the worst-group loss over the (unknown) groups in $\sG$ when $\alpha$ is close to the size of the smallest group in $\sG$.

In practice, we treat $\alpha$ as a hyperparameter.
Concretely, for some loss function $\ell(x, y; \theta)$, the CVaR objective can be written as
\begin{align}\label{eqn:jdro_objective}
    J_{\text{CVaR}}(\theta, \alpha) = \sup_{q \in \Delta^n} \Bigg \{ \sum \limits_{i=1}^n q_i \ell(x_i, y_i; \theta)~ \text{s.t.} ~\| q \|_\infty \leq \frac{1}{\alpha n} \Bigg \}, 
\end{align}
where $\Delta^n$ is the probability simplex in $\R^n$.

Note that the CVaR objective is equivalent to the average loss incurred by the $\alpha$-fraction of training points that have the highest loss.

\subsubsection{Learning from Failure (LfF)}\label{sec:lff}
LfF attempts to automatically upweight examples from challenging groups, such as those where the spurious correlation does not hold.
It does this by learning two models $f_B(y \mid x; \theta_B)$ and $f_D(y \mid x; \theta_D)$, parameterized by $\theta_B$ and $\theta_D$.

The first model $f_B$ is trained with ERM using generalized cross-entropy (GCE) loss \citep{zhang2018generalized}:
\begin{align}\label{eqn:gce}
    \ell_{\text{GCE}}(x_i, y_i; \theta_B, q) = \frac{1 - f_B(y_i \mid x_i; \theta_B)^q}{q},
\end{align}
where $q \in [0, 1)$ is a hyperparameter.
Compared to standard cross-entropy loss, the gradient of GCE loss upweights examples where $f_B(y_i \mid x_i; \theta_B)$ is large, which intentionally biases $f_B$ to perform better on easier examples and poorly on examples group challenging groups.

The second model is also trained with ERM, using cross-entropy loss, where each example $(x_i, y_i)$ is reweighted by a factor of:
\begin{align}\label{eqn:lff_weighting}
    \sW(x_i, y_i) = \frac{\log{f_B(y_i \mid x_i)}}{\log{f_B(y_i \mid x_i)} + \log{f_D(y_i \mid x_i)}}.
\end{align}
The hope is that early in training, $\log{f_B(y_i \mid x_i)}$ will be smaller than $\log{f_D(y_i \mid x_i)}$ on the easier examples, which leads to smaller weights on the easier examples and larger weights on the challenging examples.

\subsubsection{Group distributionally robust optimization (Group DRO)}\label{sec:group_dro}
Group DRO uses training group annotations to directly minimize the worst-group error on the training set. Our primary focus in this paper is the setting where we do not have access to training group annotations. However, we use group DRO as an oracle method that upper bounds the performance we can expect without any training group annotations. 
Assume we have access to group annotations on the training data such that the $n$ training points are $\{(x_1, y_1, g_1), \hdots (x_n, y_n, g_n)\}$. For some loss function $\ell(x, y; \theta)$, the group DRO objective can then be written as: 
\begin{align}\label{eqn:gdro_objective}
    J_{\text{group-DRO}}(\theta) = \max_{g \in \sG} \frac{1}{n_g} \sum_{i \mid g_i = g} \ell(x_i, y_i; \theta)
\end{align}
where $n_g$ is the number of training points with group $g_i = g$.
 \section{\ours: Just Train Twice}\label{sec:approach}

We now present \ours, a simple two-stage approach that does not require group annotations at training time.
In the first stage, we train an identification model and select examples with high training loss.
Then, in the second stage, we train a final model while upweighting the selected examples.

\paragraph{Stage 1 (identification).} 
The key empirical observation that \ours builds on is that sufficiently low complexity ERM models tend to fit groups with easy-to-learn spurious correlations (e.g., landbirds on land and waterbirds on water in the Waterbirds dataset), but not groups that do not exhibit the same correlation (e.g., waterbirds on land) \citep{sagawa2020group}.
We therefore use the simple heuristic of first training an \emph{identification model} $\pup$ via ERM and then identifying an \emph{error set} $\errors$ of training examples that $\pup$ misclassifies:
\begin{align}
\label{eqn:stepone}
     \errors = \{(x_i, y_i)~\text{s.t.}~ \pup(x_i) \neq y_i\}.
\end{align}

\paragraph{Stage 2 (upweighting).} Next, we train a final model $\pdown$ by upweighting the points in the error set $\errors$ identified in step one:
\begin{align}
\label{eqn:steptwo}
     J_{\text{up-ERM}}(\theta, \errors) = \Bigg( \upweightfactor \sum \limits_{(x, y) \in \errors} \!\!\ell(x, y; \theta) +  \sum \limits_{(x, y) \not\in \errors} \!\!\ell(x, y; \theta) \Bigg),
\end{align}
where $\upweightfactor \in \R_+$ is a hyperparameter.
The hope is that if the examples in the error sets come from challenging groups, such as those where the spurious correlation does not hold, then upweighting them will lead to better worst-group performance.

\paragraph{Practical implementation.}
Overall, training \ours is summarized in \refalg{ours}.
In practice, to restrict the capacity of the identification model, we only train it for $T$ steps, where $T$ is a hyperparameter (line 1).
This prevents it from potentially overfitting the training data and yielding an empty error set.
To implement the upweighted objective \refeqn{steptwo}, we simply upsample the examples from the error set by $\upweightfactor$ (line 3) and train the final model on the upsampled data (line 4).
Specifically, in each epoch of training, we sample each example from the error set $\upweightfactor$ times and all other examples only once.

\begin{algorithm}[t]
\begin{flushleft}
    \begin{algorithmic}
    \textbf{Input:} Training set $\sD$ and hyperparameters $T$ and $\upweightfactor$.
    
    \textbf{Stage one: identification}
    
        1. Train $\pup$ on $\sD$ via ERM for $T$ steps (Equation~\ref{eqn:erm}). 

        2. Construct the error set $\errors$ of training examples misclassified by $\pup$ (Equation~\ref{eqn:stepone}).
    
    \textbf{Stage two: upweighting identified points}
        
        3. Construct upsampled dataset $\sD_\text{up}$ containing examples in the error set $\upweightfactor$ times and all other examples once.
        
        4. Train final model $\pdown$ on $\sD_\text{up}$ via ERM (Equation~\ref{eqn:erm}).
    \end{algorithmic}
    \end{flushleft}
    \caption{\ours training
    }
    \label{alg:ours}
\end{algorithm}

We tune the algorithm hyperparameters (the number of training epochs $T$ for the identification model $\pup$, and the upweight factor $\upweightfactor$) and both identification and final model hyperparameters (e.g., the learning rate and $\ell_2$ regularization strength) based on the worst-group error of the final model $\pdown$ on the validation set.
In our experiments, we share the same hyperparameters and architecture between the identification and final models, outside of the early stopping $T$ of the identification model, and we sometimes find it helpful to learn them with different optimizers.
Note that setting the upweight factor $\upweightfactor$ to $1$ recovers ERM, so \ours should perform at least as well as ERM, given a sufficiently large validation set.
We describe full training details in \refapp{approach-details}. \begin{figure*}[ht!]
    \centering
    \includegraphics[width=\textwidth]{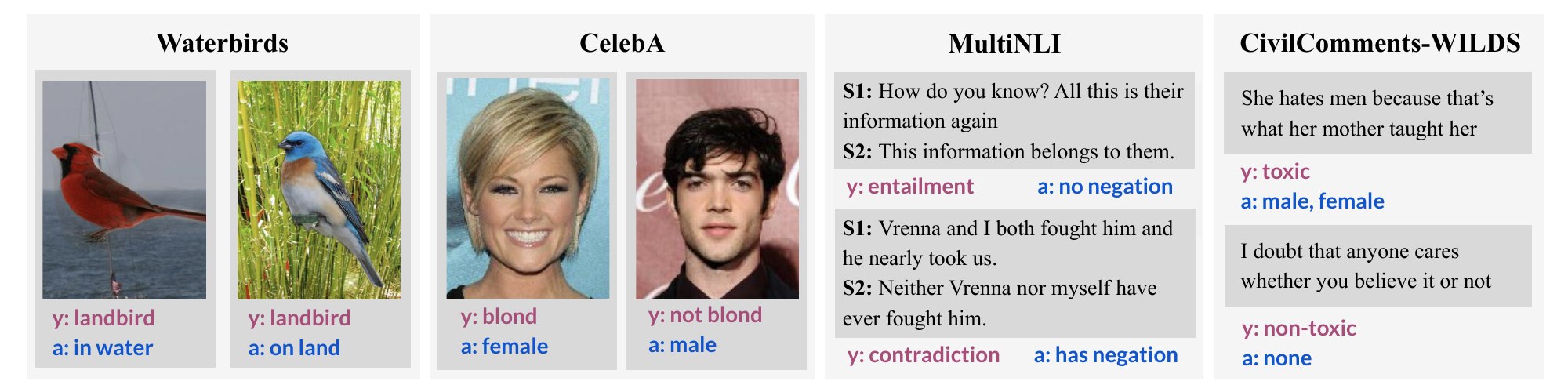}
    \vspace{-0.6cm}
    \caption{
        Examples from the tasks we evaluate on.
        The spurious attribute $\spuattribute$ is correlated with the label $y$ on the training data.
    }
    \label{fig:dataset}
\end{figure*}

\begin{table*}[t]

\label{tab:main}
    \center
    \resizebox{\textwidth}{!}{\renewcommand{\arraystretch}{1.2}
\begin{tabular}{lccccccccc}
\toprule

\multicolumn{1}{c}{Method} &
  \begin{tabular}[c]{@{}c@{}}Group labels \\in train set?\end{tabular} &
  \multicolumn{2}{c}{Waterbirds} &
  \multicolumn{2}{c}{CelebA} &
  \multicolumn{2}{c}{MultiNLI} &
  \multicolumn{2}{c}{CivilComments-WILDS} \\ 
  \cmidrule(lr){3-4} \cmidrule(lr){5-6} \cmidrule(lr){7-8} \cmidrule(lr){9-10}
 &
  \multicolumn{1}{l}{} &
  Avg Acc. &
  Worst-group Acc. &
  Avg Acc. &
  Worst-group Acc. &
  Avg Acc. &
  Worst-group Acc. &
  Avg Acc. &
  Worst-group Acc. \\ERM        & No  & 97.3\% & 72.6\% & 95.6\% & 47.2\% & 82.4\% & 67.9\% & 92.6\% & 57.4\%  \\
CVaR DRO \citep{levy2020large} & No  & 96.0\% & 75.9\% & 82.5\% & 64.4\% & 82.0\% & 68.0\% & 92.5\% & 60.5\% \\
LfF~\citep{nam2020learning} & No  &  91.2\% &  78.0\% &  85.1\% &  77.2\% & 
 80.8\% &  70.2\%
&   92.5\% &  58.8\%  \\
\ours (Ours) & No  & 93.3\% &  \textbf{86.7\%} &  88.0\% &  \textbf{81.1\%} &  78.6\% &  \textbf{72.6\%} &  91.1\% &  \textbf{69.3\%} \\
\hline
Group DRO \citep{sagawa2020group} & Yes &  93.5\% &  91.4\% &  92.9\% &  88.9\% &  81.4\% &  77.7\% &  88.9\% &  69.9\% \\
\bottomrule
\end{tabular}
}
\vspace{-0.3cm}
\caption{
        Average and worst-group test accuracies of models trained via \ours and baselines.
        \ours substantially improves worst-group accuracy relative to ERM and CVaR DRO and outperforms LfF \citep{nam2020learning}, a recently proposed algorithm for improving worst-group accuracy without group annotations.
        We also compare with group DRO, an oracle that assumes group annotations.
        \ours recovers a significant fraction of the gap in worst-group accuracy between ERM and group DRO. 
}
\end{table*}

\section{Experiments}\label{sec:experiments}

In our experiments, we first demonstrate that \ours substantially improves worst-group performance compared to standard ERM models (\refsec{mainresults}).
We also show that it recovers a significant fraction of the performance gains yielded by group DRO, which, as discussed in \refsec{preliminaries}, is an oracle that relies on group annotations on training examples.
We then present empirical analysis of \ours, including the analysis of the error set (\refsec{error_set}), exploration on the role of the validation set (\refsec{val_info}), and comparison with CVaR DRO (\refsec{cvar_dro}). 

\subsection{Setup}
We study four datasets in which prior work has observed poor worst-group performance due to spurious correlations (\reffig{dataset}). Full details about these datasets are in \refapp{dataset-details}. 

\begin{itemize}
\item \textbf{Waterbirds}~\citep{wah2011cub,sagawa2020group}: The task is to classify images of birds as ``waterbird'' or ``landbird'', and the label is spuriously correlated with the image background, which is either ``land'' or ``water.''  
\item \textbf{CelebA}~\citep{liu2015deep}: We consider the task from \citet{sagawa2020group} of classifying the hair color of celebrities as ``blond'' or ``not blond.''
The label is spuriously correlated with gender, which is either ``male'' or ``female.''
\item \textbf{MultiNLI}~\citep{williams2018broad}: Given a pair of sentences, the task is to classify whether the second sentence is entailed by, neutral with, or contradicts the first sentence.
We use the spurious attribute from \citet{sagawa2020group}, which is the presence of negation words in the second sentence; due to the artifacts from the data collection process, contradiction examples often include negation words.

\item \textbf{CivilComments-WILDS}~\citep{borkan2019nuanced,koh2021wilds}: The task is to classify whether an online comment is toxic or non-toxic, and the label is spuriously correlated with mentions of certain demographic identities (male, female, White, Black, LGBTQ, Muslim, Christian, and other religion).
We use the evaluation metric from \citet{koh2021wilds}, which defines 16 overlapping groups $(a, \emph{toxic})$ and $(a, \emph{non-toxic})$ for each of the above 8 demographic identities $a$, and report the worst-group performance over these groups.
\end{itemize}

\paragraph{Points of comparison.}
We aim to answer two main questions:
(1) How does \ours compare with other approaches that also do not use training group information?
(2) How does \ours compare with approaches that \emph{do} use training group information?

To answer the first question, we compare \ours with ERM, CVaR DRO, and a recently proposed approach called Learning from Failure (LfF) \citep{nam2020learning}.
To answer the second question, we compare \ours with group DRO \citep{sagawa2020group}, an oracle that uses training group annotations.
For details about these approaches, see \refsec{preliminaries}.
Note that on CivilComments, group DRO cannot be directly applied on the 16 defined groups, since it is not designed for overlapping groups.
Instead, our group DRO minimizes worst-group loss over 4 groups $(y, a)$, where the spurious attribute $a$ is a binary indicator of whether any demographic identity is mentioned and the label $y$ is \emph{toxic} or \emph{non-toxic}.
We tune the hyperparameters of all approaches based on worst-group performance on a small validation set with group annotations.

\subsection{Main Results}
\label{sec:mainresults}

\reftab{main} reports the average and worst-group accuracies of all approaches.
Compared to other approaches that do not use training group information, \ours consistently achieves higher worst-group accuracy on all 4 datasets.
Additionally, \ours performs well even relative to approaches that use training group information.
In particular, \ours recovers a significant portion of the gap in worst-group accuracy between ERM and group DRO, closing 75\% of the gap on average.
As a caveat, we note that simple label balancing also achieves comparable worst-group accuracy to group DRO on CivilComments.

\ours's worst-group accuracy improvements come at only a modest drop in average accuracy, averaging only 4.2\% worse than the highest average accuracy on each dataset.
This drop is consistent with \citet{sagawa2020group}, which observes a tradeoff between average and worst-group accuracies.

\begin{table}[t]
\centering
\resizebox{\columnwidth}{!}{\begin{tabular}{cccc}
\toprule
\begin{tabular}[c]{@{}c@{}}Dataset\end{tabular} &
  \begin{tabular}[c]{@{}c@{}}Worst-group\\Recall\end{tabular} &
  \begin{tabular}[c]{@{}c@{}}Worst-group\\Precision\end{tabular} &
  \begin{tabular}[c]{@{}c@{}}Worst-group\\Empirical Rate\end{tabular} \\
  \cmidrule(lr){1-1} \cmidrule(lr){2-2} \cmidrule(lr){3-3} \cmidrule(lr){4-4}
Waterbirds & 87.5\% & 19.1\% & 1.2\% \\
CelebA     & 94.7\%  & 9.4\% & 0.9\% \\
MultiNLI   & 67.1\% & 2.2\% & 1.0\% \\
CivilComments & 96.9\% & 7.8\% & 0.9\% \\
\bottomrule
\end{tabular}
}
\center\small
\vspace{-0.3cm}
    \caption{
        The precision and recall of the worst-group examples (i.e., the group with lowest validation accuracy) belonging to \ours's error set.
        The error set includes a high fraction of the worst-group examples and includes them at a much higher rate than they occur in the training data.
}
\label{tab:recall-precision}
\end{table}

\begin{table}[t]
\centering
\resizebox{\columnwidth}{!}{\begin{tabular}{ccc}
\toprule
Group & Enrichment & ERM test acc. \\
  \cmidrule(lr){1-1} \cmidrule(lr){2-2} \cmidrule(lr){3-3}
(land background, waterbird) & 15.92x & 72.6\% \\
(water background, landbird) & 6.97x & 73.3\% \\
(water background, waterbird) & 2.40x & 96.3\% \\
(land background, landbird) & 0.02x & 99.3\% \\
\bottomrule
\end{tabular}
}
\center\small
\vspace{-0.3cm}
    \caption{Waterbirds error set breakdowns.}
\label{tab:waterbirds-breakdown}
\end{table}

\begin{table}[t]
\centering
\begin{tabular}{ccc}
\toprule
Group & Enrichment & ERM test acc. \\
  \cmidrule(lr){1-1} \cmidrule(lr){2-2} \cmidrule(lr){3-3}
(male, blond) & 10.44x & 47.2\% \\
(female, blond) & 5.42x & 89.1\% \\
(male, non-blond) & 0.32x & 99.3\% \\
(female, non-blond) & 0.01x & 95.1\% \\
\bottomrule
\end{tabular}
\center\small
\vspace{-0.3cm}
    \caption{CelebA error set breakdowns.
    }
\label{tab:celeba-breakdown}
\end{table}

\begin{table}[!h]
\centering
\begin{tabular}{ccc}
\toprule
Group & Enrichment & ERM test acc. \\
  \cmidrule(lr){1-1} \cmidrule(lr){2-2} \cmidrule(lr){3-3}
(negation, neutral) & 2.2x & 67.9\% \\
(no negation, contradiction) & 1.35x & 77.0\% \\
(negation, entailment) & 1.14x & 80.4\% \\
(no negation, neutral) & 1.07x & 81.8\% \\
(no negation, entailment) & 0.73x & 86.1\% \\
(negation, contradiction) & 0.19x & 94.5\% \\
\bottomrule
\end{tabular}
\center\small
\vspace{-0.3cm}
    \caption{MultiNLI error set breakdowns.}
\label{tab:multinli-breakdown}
\end{table}

\subsection{Error set analysis}\label{sec:error_set}
We find it surprising that just a small amount of group information on the validation set can allow \ours to achieve high worst-group accuracy with no knowledge of the groups on the training set.
We now probe into how \ours achieves such high worst-group accuracy.
In order to perform this analysis, we use the group annotations on the training data to closely examine what examples are upweighted in the error set identified in the first step of \ours, though we don't use such training group annotations for training \ours.

To start, we define the worst group as the group on which the standard ERM model achieves the lowest test accuracy, when tuned for worst-group validation accuracy.
We analyze how well the error set captures this worst group.
To do this, we measure \emph{precision}, the fraction of examples in the error set that belong to the worst group, \emph{recall}, the fraction of the worst group examples that are included in the error set, and the \emph{empirical rate}, the rate at which the worst group examples appear in the training data.

As reported in \reftab{recall-precision}, we observe that the error set contains worst-group examples at a much higher rate (precision) than they appear in the training dataset (empirical rate).
Worst-group examples appear in the error set 2.2x to 15.9x more frequently in the error set than in the training data, across the 4 datasets.
In other words, the worst group is significantly \emph{enriched} in the error set compared to the training dataset, which may explain why \ours has much better worst-group performance over ERM.
Additionally, the error set has high worst-group recall, ranging from 67.1\% to 96.9\% and averaging to 86.4\% across the 4 datasets.
Together, these results indicate that the worst group is included in the error set at relatively high both precision and recall.

Empirically, ERM performs poorly on several groups, not just on a single worst group.
We therefore next examine what other groups the examples in the error set belong to, beyond the worst group.
For each group, we compute two metrics:
(i) \emph{enrichment}, defined as how much more frequently examples from a group appear in the error set than in the training data (i.e., the precision of the group divided by the empirical rate of the group);
(ii) the test accuracy that ERM achieves on this group, when tuned for worst-group validation accuracy.

Tables~\ref{tab:waterbirds-breakdown} to \ref{tab:multinli-breakdown} and \reftab{civilcomments-breakdown} in \refapp{cc_breakdown} report these results for Waterbirds, CelebA, MultiNLI, and CivilComments respectively.
We observe that the enrichment roughly inversely correlates with ERM's test accuracy on that group:
examples from low performance groups are included at high rates in the error set relative to the empirical rate.
This may help \ours perform better across all groups that ERM performs poorly on, which in turn improves worst-group accuracy.

\begin{table}[t]
\centering
\resizebox{\columnwidth}{!}{\begin{tabular}{cc}
\toprule
& Worst-group test acc. \\
\cmidrule(lr){2-2}
Standard error set & 86.7\% \\
No waterbirds on water backgrounds & 80.7\% \\
Swap error set examples & 86\% \\
\bottomrule
\end{tabular}
}
\center\small
    \caption{Worst-group test accuracies with 3 variants of the error set on Waterbirds: (i) standard unchanged error set; (ii) removing all waterbird on water background examples; (iii) swapping each error set example with a random example from the same group.
    }
\label{tab:waterbirds_misc}
\end{table}

Finally, we note that while the groups with high enrichment often correspond to groups where the spurious correlation does not hold, this is not always the case.
In particular, the waterbird on water background group in Waterbirds and the blonde female group in CelebA have high enrichments, even though the spurious correlation holds in these groups.
We hypothesize that this occurs due to label imbalance, since the waterbird label and blonde label are relatively rare and appear only in 23\% and 15\% of the training examples, respectively.
Empirically, upweighting examples from these groups is indeed important for \ours's worst-group test accuracy.
When we remove all waterbird on water background examples from the error set, \ours's worst-group test accuracy drops by 6\%.
However, while the group composition of the error set (i.e., the fraction of the error set in each group) is important, the exact examples inside the error set do not seem to matter.
When we replace each error set example with another example from the same group on Waterbirds, worst-group test accuracy drops by only 0.7\%.
These two results are shown in \reftab{waterbirds_misc}.

\subsection{Hyperparameter Tuning and the Role of the Validation Set}\label{sec:val_info}

\begin{table*}[t]
\centering
\resizebox{\linewidth}{!}{\begin{tabular}{lcccc}
\toprule
           & \multicolumn{2}{c}{Waterbirds worst-group test acc.}        & \multicolumn{2}{c}{CelebA worst-group test acc.}            \\
\cmidrule(lr){2-3} \cmidrule(lr){4-5}
           & Tuned for average & Tuned for worst-group & Tuned for average & Tuned for worst-group \\
CVaR DRO \citep{levy2020large}    & 62.0\%        & 75.9\%                & 36.1\%        & 64.4\%                \\
LfF \citep{nam2020learning}       & 44.1\%        & 78.0\%                & 24.4\%        & 77.2\%                  \\
JTT (Ours)                        & 62.5\%        & 86.7\%                & 40.6\%        & 81.1\%         \\      
\bottomrule
\end{tabular}}
\caption{Across the methods that do not use training group annotations, worst-group test performance is significantly higher when hyperparameters are tuned for worst-group validation accuracy instead of average validation accuracy. This shows that for these methods, it is still critical to have validation group annotations.}
\label{tab:tuning}
\end{table*}

In all of our experiments, we tune the algorithm and model hyperparameters based on the worst-group accuracy on the validation set. In general, across all methods, we found hyperparameter tuning in this fashion to be critical. \reftab{tuning} shows that even for CVaR DRO, LfF, and \ours, which all try to improve worst-group accuracy without relying on training group annotations, the worst-group test accuracies on Waterbirds and CelebA plummet when the hyperparameters are tuned for average accuracy on the validation set, instead of worst-group accuracy on the validation set. Importantly, this means that even though these methods do not require training group annotations, they still require \emph{validation group annotations} in order to have high worst-group test accuracy. Existing methods for improving worst-group accuracy generally rely on some form of this assumption (either by explicitly tuning for validation group accuracy, or by assuming access to a validation set that is balanced by groups). Removing this reliance is an important direction for future work; we discuss this further in \refsec{discussion}.

The sensitivity of CVaR DRO, LfF, and \ours to hyperparameters is consistent with prior work showing that reweighting methods like these require appropriate capacity control via techniques like early stopping or strong $\ell_2$ regularization \citep{byrd2019effect, sagawa2020group}. These parameters---e.g., how many epochs to train for, or what $\ell_2$ regularization strength to use---generally need to be set by tuning directly for worst-group accuracy; we found that tuning for average accuracy typically results in models with lower $\ell_2$ regularization strength and that were trained for more epochs.

\begin{figure}
    \centering
    \includegraphics[width=0.9\columnwidth]{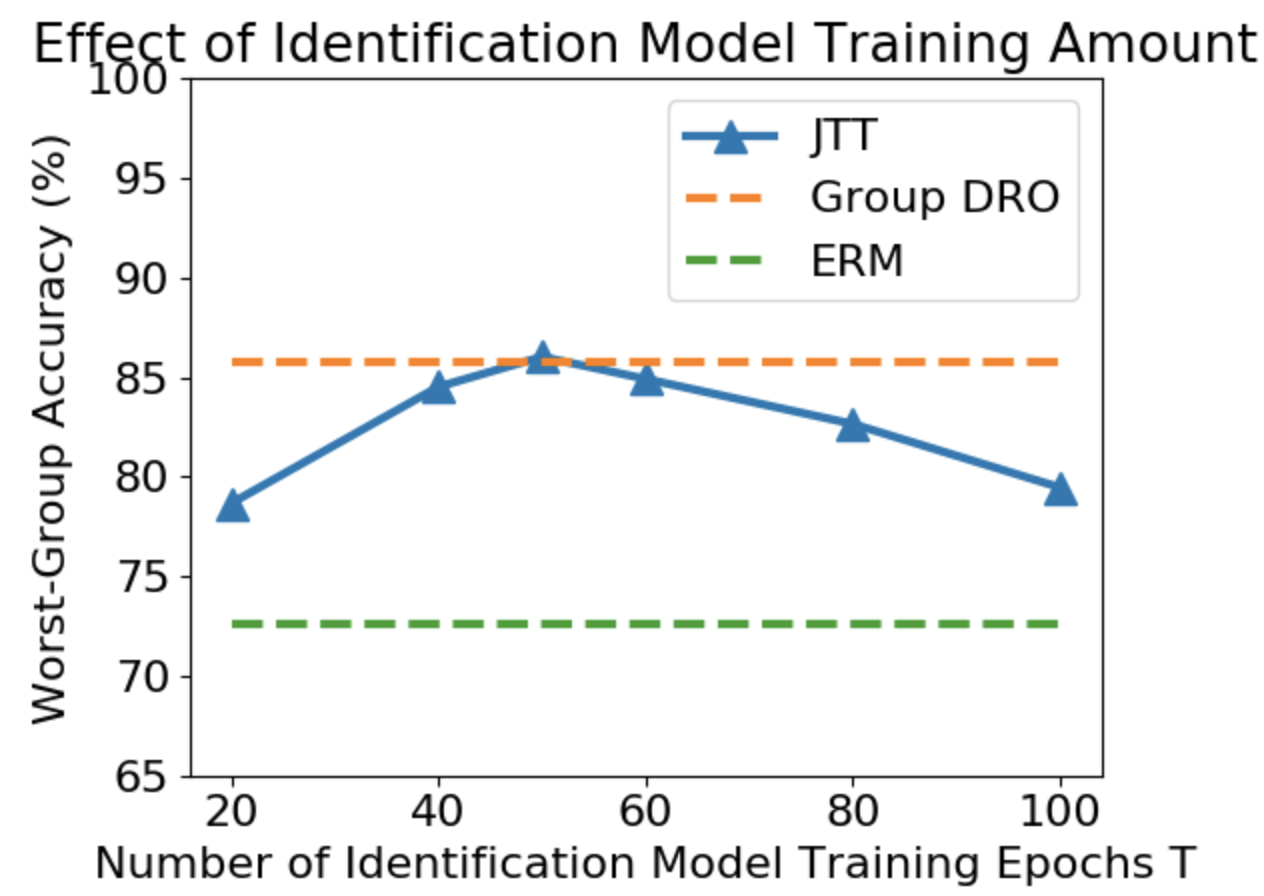}
    \vspace{-0.1cm}
    \caption{
        Effect of number of epochs of identification model training in \ours on Waterbirds.
        Worst-group test accuracy is high for $T$ between 40 and 60 epochs, but degrades when $T$ becomes too small or large, which results in less informative error sets.
    }
    \label{fig:timing-ablation}
\end{figure}

Compared to ERM, \ours has two additional hyperparameters: the number of epochs to train the identification model $T$ and the upweight factor $\upweightfactor$.
As an illustration of the sensitivity to hyperparameters, \reffig{timing-ablation} shows how the worst-group accuracy of \ours's final model changes as we vary $T$ between 20 and 100 epochs on Waterbirds.
Worst-group accuracy is high when $T$ is between 40 and 60, but drops when $T$ is too small or too large.

\begin{table}
\centering
\footnotesize
\begin{tabular}{cccccccc}
\toprule
\multicolumn{4}{c}{Waterbirds} & \multicolumn{4}{c}{CelebA} \\
1x & $\frac{1}{5}$x & $\frac{1}{10}$x & $\frac{1}{20}$x & 1x & $\frac{1}{5}$x & $\frac{1}{10}$x & $\frac{1}{20}$x \\
\cmidrule(lr){1-4} \cmidrule(lr){5-8}
\!\!\textbf{86.7\%}\!\! & \textbf{84.0\%}\!\! & \!\!\textbf{86.9\%}\!\! & \!\!76.0\%\!\! & \!\!\textbf{81.1\%}\!\! & \!\!\textbf{81.1\%}\!\! & \!\!\textbf{81.1\%}\!\! & \!\!\textbf{82.2}\%\!\! \\
\bottomrule
\end{tabular}
\center
\caption{
\ours retains high test worst-group accuracy on Waterbirds and CelebA when the validation set size is reduced to $\frac{1}{10}$x, though performance drops at $\frac{1}{20}$x on CelebA.
}
\label{tab:val_set_size}
\vspace{-5mm}
\end{table}

\paragraph{Reducing the size of the validation set.}
In the main experiments in \refsec{mainresults}, we use the default validation sets provided with each of the datasets.
Using group annotations on these validation sets is already cheaper than training group annotations, as these default validation sets are 2--10x smaller than their corresponding training sets.
However, we additionally test if \ours can continue to achieve high worst-group performance using even smaller validation sets to further reduce the cost of obtaining group annotations on those sets.
On Waterbirds and CelebA, we reduce the validation set size by a factor of 1x (no reduction), $\frac{1}{5}$x, $\frac{1}{10}$x, and $\frac{1}{20}$x and tune \ours's hyperparameters based on worst-group accuracy on the reduced validation set.
We find that \ours continues to achieve high worst-group accuracy, even when reducing the validation set size by $\frac{1}{10}$x and $\frac{1}{20}$x, amounting to only 119 and 993 total examples, on Waterbirds and CelebA respectively.

\subsection{Comparison with CVaR DRO}
\label{sec:cvar_dro}

In this section, we explore the relation between \ours and CVaR DRO.
Recall that the CVaR objective in Equation~\ref{eqn:jdro_objective} is the average loss incurred by the $\alpha$-fraction of training examples with the highest loss. 
We can view minimizing this objective as upweighting this $\alpha$-fraction of examples while ignoring the remaining examples. 
In this way, \ours is conceptually similar to CVaR DRO: both upweight training points with high loss, without requiring group annotations of training points. However, their empirical performance is widely different: CVaR DRO offers only small gains in worst-group accuracy over ERM, while \ours offers substantial gains.
One key difference is that in \ours, the set of points that get upweighted $\errors$ is computed once during stage 1, and then held fixed. 
In contrast, minimizing the CVaR objective involves dynamically computing the $\alpha$-subset of points with the highest loss at each step, upweighting them and updating the model, and then repeating to update the $\alpha$-subset. 
As we show next, ablating this key difference from \ours substantially degrades worst-group performance.

\paragraph{Dynamically computing the error set in \ours lowers accuracy.}
We start by observing that the performance of \ours drops when we dynamically recompute the error set $\errors$, instead of only computing $\errors$ once using the identification model.
Concretely, we study a variant of \ours on the Waterbirds dataset: as usual, we first train an identification model for $T=50$ epochs, but then while training the final model, every $K$ epochs, we dynamically update the error set $\errors$ as the errors of the final model over the training set.
Setting $K$ to be $\infty$---which means that we only compute the error set $\errors$ once after training the identification model for $T$ epochs---recovers standard \ours.
On the other hand, lowering $K$ makes the algorithm more similar to minimizing CVaR, since this more frequently updates the upweighted set to be the examples with higher loss under the current model, instead of the examples with higher loss under the static identification model.

\reftab{ablation-freezing} shows the results as we vary $K$ between $10$, $20$, $30$, and $50$ epochs on Waterbirds,
re-tuning all hyperparameters for each value of $K$.
At high values of $K$, where the error set remains fixed for many epochs, both average and worst-group accuracies are high.
However, as $K$ decreases, the average and worst-group accuracies drop.
These results show that at least on Waterbirds, holding the error set fixed appears to be critical for \ours.

\begin{table}
\centering
\small
\begin{tabular}{ccc}
\toprule
 & \multicolumn{2}{c}{Waterbirds}    \\
            \cmidrule(lr){1-1} \cmidrule(lr){2-3}
Epochs per update ($K$) & Average Acc. & Worst-group Acc. \\
10          & 89.2\%      & 77.1\%              \\
20          & 86.3\%      & 72.1\%              \\
30          & 91.4\%      & 86.8\%              \\
50          & 93.1\%      & 88.6\%              \\
\bottomrule
\end{tabular}
\center\small
    \caption{
        Effect of dynamically computing \ours's error set on Waterbirds.
        We first train the identification model for $T=50$ epochs, as usual. Then, we dynamically update the error set using the final model after every $K$ epochs of training the final model.
        Lower values of $K$ have significantly lower accuracies.
    }
\label{tab:ablation-freezing}
\end{table}

\begin{figure*}[t]
    \centering
    \includegraphics[width=\textwidth]{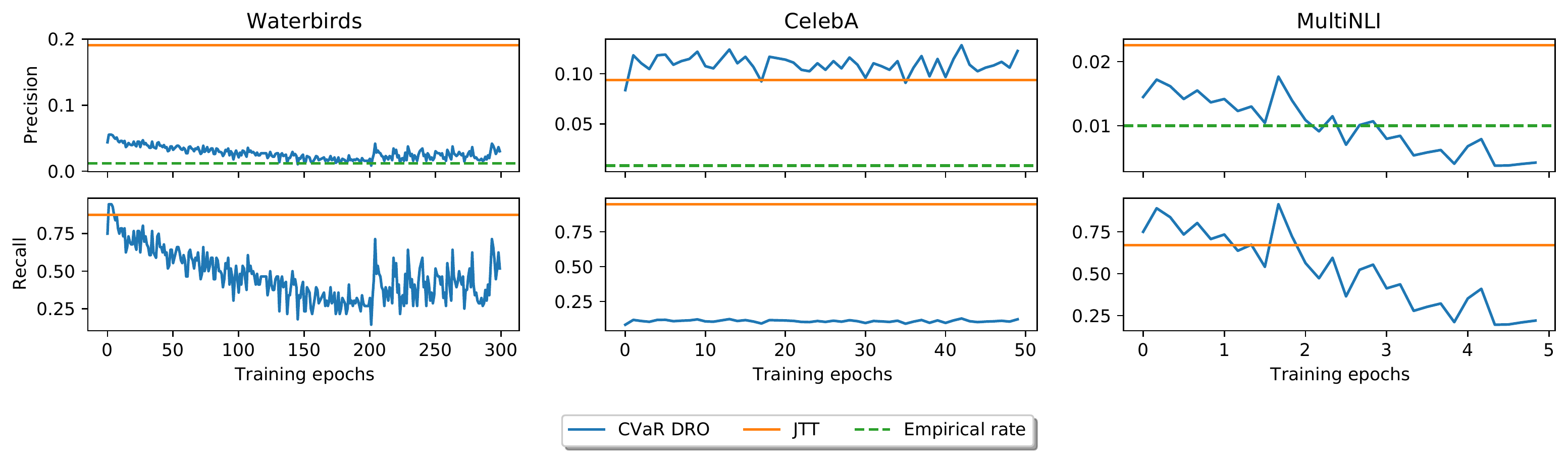}
    \vspace{-0.3cm}
    \caption{
    The composition of the CVaR set (the $\alpha$-fraction of training examples with the highest loss) as training progresses for CVaR DRO models. 
    In these plots, the worst group is defined as the group with the lowest test accuracy under the ERM model.
    For each dataset, the top plot shows the worst-group precision: the fraction of the CVaR set that belongs to the worst group (blue), with the analogous fraction of the \ours error set (orange) and the overall training data (green) provided for comparison.
    The bottom plot shows the worst-group recall: the fraction of total worst-group examples that are in the respective sets.
    For Waterbirds and MultiNLI, the CVaR set is less enriched for the worst group compared to \ours.
    For CelebA, it is slightly more enriched, but $\alpha$ is much smaller than the size of the \ours error set, so it only upweights a small fraction of the worst group.
    }
    \label{fig:joint_dro_points}
\end{figure*}

\paragraph{Which examples does CVaR upweight?}
The analysis above suggests that the relatively poor worst-group performance of CVaR DRO might stem from how it dynamically computes which examples to upweight. 
We further study the behavior of CVaR DRO by analyzing the examples that it upweights.
Concretely, throughout CVaR DRO training, we periodically identify the $\alpha$-fraction of training examples with the highest loss and measure the worst-group precision and recall, where the worst group is defined as the group on which ERM achieves the lowest test accuracy.

\reffig{joint_dro_points} shows the results using the value of $\alpha$ achieving the highest worst-group validation accuracy: $\alpha = 0.2$ on Waterbirds, $\alpha = 0.00852$ on CelebA, and $\alpha = 0.5$ on MultiNLI.
On Waterbirds, the worst-group examples (which comprise approximately 1\% of the training set) make up 19\% of the error set for \ours, whereas they oscillate between 1\% and 10\% of the worst-$\alpha$ fraction for CVaR DRO. 
As a result, \ours consistently upweights nearly 90\% of the worst-group examples, whereas CVaR DRO oscillates between upweighting the worst group and the other groups, upweighting as little as 20\% of the examples at some points during training.
On CelebA, CVaR DRO upweights the worst-group examples with slightly higher precision than \ours, but $\alpha$ is much smaller than the size of the error set; as a result, \ours upweights nearly 95\% of the worst-group examples, whereas CVaR DRO only upweights 13\% of them. 
On MultiNLI, the worst group steadily gets less and less upweighted for CVaR DRO, whereas \ours upweights it at a higher rate, though it still only comprises 2\% of the error set for \ours. 

These results suggest that the CVaR objective might be overly conservative where the $\alpha$-fraction of examples with highest loss often include many examples from other groups. Furthermore, the set of examples varies widely across different iterations of training. In contrast, \ours upweights a fixed set of points. Empirically, we find that this allows \ours to successfully use the worst-group accuracy on a small validation set to identify error sets that improve accuracy on groups we care about.

 \section{Discussion}\label{sec:discussion}

In this work, we presented Just Train Twice (\ours), a simple algorithm that substantially improves worst-group performance without requiring expensive group annotations during training.
We conclude by discussing several directions for future work.

First, a better theoretical understanding of when and why \ours works would help us to refine and further develop methods for training models that are less susceptible to spurious correlations. For example, it would be useful to understand why early-stopped ERM models (as in the identification models used by \ours) seem to consistently latch onto the spurious correlations in our datasets, and why it seems to be important to fix the upweighted set instead of dynamically recomputing it, as in CVaR DRO.

Second, \ours and many prior methods on robustness without group information all rely on a validation set that is representative of the distribution shift or annotated with group information. 
While these annotations are significantly cheaper that labeling the entire training set, it still requires the practitioner to be aware of any spurious correlations and define groups accordingly. 
Doing so may be notably difficult in real-world applications. 
Therefore, this leaves open the question of whether methods can perform well with mis-specified groups or no group annotations whatsoever. 

Finally, while our experiments focus on group robustness in the presence of spurious correlations, \ours is not specifically tailored to spurious correlations. Given \ours's simplicity, it would be straightforward to experiment with \ours to see if it might improve performance under different types of distribution shifts, such as in domain generalization settings~\citep{blanchard2011generalizing, muandet2013domain}.

\paragraph{Reproducibility.} Our code is publicly available at \url{https://github.com/anniesch/jtt}. 
\section*{Acknowledgements}

This work was supported by NSF Award Grant No.~1805310 and in part by Google.
EL is supported by a National Science Foundation Graduate Research Fellowship under Grant No.~DGE-1656518.
AR is supported by a Google PhD Fellowship and Open Philanthropy Project AI Fellowship.
SS is supported by a Herbert Kunzel Stanford Graduate Fellowship.

\bibliography{all}
\bibliographystyle{icml2021}

\newpage
\clearpage
\appendix
\section{Training Details}\label{sec:approach-details}
In this section, we detail the model architectures and hyperparameters used by each approach.
Within each dataset, we used the same model architecture across all approaches:
ResNet-50 \citep{he2016resnet} for Waterbirds and CelebA, and BERT for MultiNLI and CivilComments \citep{devlin2019bert}. 
For ResNet-50, we used the PyTorch \citep{paszke2017automatic} implementation of ResNet-50, starting from ImageNet-pretrained weights. 
For BERT, we used the the HuggingFace implementation \citep{wolf2019transformers} of BERT, also starting from pretrained weights.

We use the LfF implementation released by \citet{nam2020learning}.
We use the group DRO and ERM implementations released by \citet{sagawa2020group} and also implement CVaR DRO and \ours on top of this code base, with the CVaR DRO implementation adapted from \citet{levy2020large}.
For the group DRO experiments on Waterbirds, CelebA, and MultiNLI, we directly use the reported performance numbers from \citet{sagawa2020group}. We note that these numbers utilize group-specific loss adjustments that encourage the model to attain lower training losses on smaller groups, which was shown to improve worst-group generalization. We train our own group DRO model on CivilComments-WILDS as it was not included in \citet{sagawa2020group}; for this, we did not implement these group adjustments.
We train our own models for all other algorithms. 

For all approaches, we tune all hyperparameters as well as early stop based on highest worst-group accuracy on the validation set. 
On top of the hyperparameters shared between all algorithms (e.g., learning rate, $\ell_2$ regularization), which we detail for each dataset below, CVaR DRO, \ours, and LfF have additional hyperparameters that we tuned separately for each dataset:
\begin{itemize}
    \item For \textbf{CVaR DRO}, we tune the size of the worst-case subpopulation $\alpha \in \{0.1, 0.2, 0.5\}$. For CelebA, we additionally tried $\alpha = \frac{\text{\# smallest group examples}}{\text{\# training examples}} = 0.00852$.
\item For \textbf{LfF}, we tune the hyperparameter $q$ by grid searching over $q \in \{0.1, 0.3, 0.5, 0.7, 0.9\}$. For CivilComments, we additionally sample two values log-uniformly from $(0, 0.1]$.
    This hyperparameter was not tuned in the experiments in \citet{nam2020learning}.
    \item For \textbf{\ours}, we additionally tune the number of epochs of training the identification model $T$ and the upsampling factor $\upweightfactor$.
    While developing \ours, we tried the following values of $T$ and $\upweightfactor$ without looking at test results, though not necessarily all combinations:
    $T \in \{1, 2, 40, 50, 60\}$ and $\upweightfactor \in \{5, 10, 20, 30, 40, 50, 100, \frac{|\text{training set}|}{|\text{error set}|}\}$.
    For the final experiments we tune over $\upweightfactor \in \{20, 50, 100\}$ for the vision datasets (Waterbirds and CelebA) and $\upweightfactor \in \{4, 5, 6\}$ for the NLP datasets (MultiNLI and CivilComments).
    Additionally, Waterbirds requires more training epochs than the others, due to its much smaller training set size, so we tune over $T \in \{40, 50, 60\}$ for Waterbirds and $T \in \{1, 2\}$ for all other datasets.
\end{itemize}
ERM has no additional algorithm-specific hyperparameters.
For group DRO, we fixed the step size $\eta_q$ for updating group weights to its default value of $0.01$ from \citet{sagawa2020group}, without tuning.

In general, for \ours, we fixed the initialization model and the final model to share the same hyperparameters, with two exceptions.
First, the initialization model is trained only for $T$ epochs, whereas the final model is trained for longer; exact values vary by dataset.
Second, for BERT, we found that it was helpful for \ours to be able to choose different optimizers for the initialization model and final model.
Specifically, for the ResNet-50 models, we used SGD with momentum 0.9 and no learning rate scheduler or gradient clipping. For BERT, we additionally considered the standard AdamW optimizer \citet{loshchilov2017sgdr} with a linearly-decaying learning rate and gradient clipping (setting the max $\ell_2$-norm of the gradients to 1), but we set this as a hyperparameter and allowed the initialization and final models to independently be optimized by SGD or AdamW.

\begin{table}
\centering
\small
\begin{tabular}{cc}
\toprule
Learning rate & $\ell_2$ regularization strength \\
1e-3 & 1e-4 \\
1e-4 & 1e-1 \\
1e-5 & 1 \\
\bottomrule
\end{tabular}
\center
    \caption{
        Learning rates and $\ell_2$ regularization strengths for Waterbirds.
    }
\label{tab:waterbirds_lr}
\end{table}

\begin{table}
\centering
\small
\begin{tabular}{cc}
\toprule
Learning rate & $\ell_2$ regularization strength \\
1e-4 & 1e-4 \\
1e-4 & 1e-2 \\
1e-5 & 1e-1 \\
\bottomrule
\end{tabular}
\center
    \caption{
        Learning rates and $\ell_2$ regularization strengths for CelebA.
    }
\label{tab:celeba_lr}
\end{table}

\paragraph{Waterbirds.}
All approaches are optimized for up to $300$ epochs with batch size $64$,
using batch normalization \citep{ioffe2015batch}, and no data augmentation.
We chose this smaller batch size (compared to the batch size of 128 used in \citet{sagawa2020group}) for computational convenience.
All approaches are optimized with stochastic gradient descent (SGD) with momentum $0.9$.

For all approaches, we tune over the 3 pairs of learning rate and $\ell_2$ regularization strength used by \citet{sagawa2020group} detailed in \reftab{waterbirds_lr}.
For ERM and LfF, this yields learning rate 1e-3 and $\ell_2$ regularization 1e-4.
For CVaR DRO, this yields learning rate 1e-4 and $\ell_2$ regularization 1e-1.
For \ours and group DRO, this yields learning rate 1e-5 and $\ell_2$ regularization 1.

Our grid search over $\alpha$ for CVaR DRO yields $\alpha = 0.2$.
Our grid search over $q$ for LfF yields $q = 0.5$.
Our grid search over $T$ and $\upweightfactor$ for \ours yields $T = 60$ epochs and $\upweightfactor = 100$.

\paragraph{CelebA.}

We train all approaches for up to $50$ epochs with batch size $128$, using batch normalization and no data augmentation.
Like in Waterbirds, we optimize all approaches with SGD with momentum $0.9$.

Also like in Waterbirds, we tune over the 3 pairs of learning rate and $\ell_2$ regularization strength used by \citet{sagawa2020group}, detailed in \reftab{celeba_lr}.
For ERM, this yields learning rate 1e-4 and $\ell_2$ regularization 1e-4.
For LfF, this yields learning rate 1e-4 and $\ell_2$ regularization 1e-2.
For all other approaches, this yields learning rate 1e-5 and $\ell_2$ regularization 1e-1.

Our grid search over $\alpha$ for CVaR DRO yields $\alpha = 0.00852$.
Our grid search over $q$ for LfF yields $q = 0.5$.
Our grid search over $T$ and $\upweightfactor$ for \ours yields $T = 1$ epoch and $\upweightfactor = 50$.

\paragraph{MultiNLI.}
We train each approach for up to $5$ epochs with default tokenization, dropout, batch size $32$, no $\ell_2$-regularization, and an initial learning rate of $0.00002$.

\ours achieves the highest validation worst-group accuracy using SGD optimization without clipping for the initial model, and using the AdamW optimizer with clipping for the final model.
All other approaches achieve highest validation worst-group accuracy using AdamW with clipping.
Our grid search over $\alpha$ for CVaR DRO yields $\alpha = 0.5$.
Our grid search over $q$ for LfF yields $q = 0.1$.
Our grid search over $T$ and $\upweightfactor$ for \ours yields $T = 2$ epochs and $\upweightfactor = 6$.

\paragraph{CivilComments-WILDS.}

All approaches use the details from \citet{koh2021wilds}.
We capped the number of tokens per example at $300$ and used an initial learning rate of $0.00001$.
We train all approaches for up to $5$ epochs with batch size $16$ and $\ell_2$ regularization strength of $0.01$.

\ours achieves the highest validation worst-group accuracy using SGD optimization without clipping for the initial model, and using the AdamW optimizer with clipping for the final model.
All other approaches achieve highest validation worst-group accuracy using AdamW with clipping.
Our grid search over $\alpha$ for CVaR DRO yielded $\alpha = 0.5$.
Our grid search over $q$ for LfF yielded $q = 0.00001$.
Our grid search over $T$ and $\upweightfactor$ for \ours yields $T = 2$ epochs and $\upweightfactor = 6$.

We also note that our group DRO approach uses a different spurious attribute compared to the group DRO results reported in \citet{koh2021wilds}.
Our group DRO uses the spurious attribute of any demographic identity being mentioned, while the one in \citet{koh2021wilds} uses only mentions of the Black demographic.
Both perform similarly: ours achieves 0.3\% lower worst-group accuracy, but 0.5\% higher average accuracy.

\section{Dataset Details}\label{sec:dataset-details}
\subsection{Waterbirds}
We use the Waterbirds dataset introduced by \citet{sagawa2020group}, which is constructed by cropping out images of birds from the CUB dataset \citep{wah2011cub} and pasting them on backgrounds from the Places dataset \citep{zhou2017places}.
In this dataset, images of seabirds (albatross, auklet, cormorant, frigatebird, fulmar, gull, jaeger, kittiwake, pelican, puffin, or tern) and waterfowl (gadwall, grebe, mallard,
merganser, guillemot, or Pacific loon) are labeled as \emph{waterbirds}, and all other birds are labeled as \emph{landbirds}.

Backgrounds from the \emph{ocean} and \emph{natural lake} categories in the Places dataset are considered to have spurious attribute $\spuattribute =$ \emph{water background}, while backgrounds from the \emph{bamboo forest} or \emph{broadleaf forest} categories are considered to have spurious attribute $\spuattribute =$ \emph{land background}.

There are two minority groups: (land background, waterbird) and (water background, landbird); and two majority groups: (land background, landbird) and (water background, waterbird).
We use the same training / valid / test splits from \citet{sagawa2020group}.
In the training data, 95\% of the waterbirds appear on water backgrounds, and 95\% of the landbirds appear on land backgrounds, so the minority groups contain far fewer examples than the majority groups.
In the validation and test sets, both the landbirds and waterbirds are evenly split between the water and land backgrounds.

\subsection{CelebA}

We use the task setup from \citet{sagawa2020group} on the CelebA celebrity face dataset \citep{liu2015deep}.
The label $y$ is set to be the \emph{Blond\_Hair} attribute, and the spurious attribute $\spuattribute$ is set to be the \emph{Male} attribute: being female spurious correlates with having blond hair.
The minority groups are (blond, male) and (not blond, female), although the (blond, male) group is significantly smaller than the (not blond, female) group.
The majority groups are (blond, female) and (not blond, male).
We use the standard train / valid / test splits from \citet{sagawa2020group}.

\subsection{MultiNLI}

We use the task setup from \citet{sagawa2020group} on the MultiNLI natural language inference dataset \citep{williams2018broad}.
Given two sentences, a premise and a hypothesis, the task is to predict whether the hypothesis is \emph{entailed by}, \emph{neutral with}, or \emph{contradicted by} the premise.
The spurious attribute $\spuattribute$ is a binary indicator for when any of the negation words \emph{nobody}, \emph{no}, \emph{never}, or \emph{nothing} appear in the second sentence (the hypothesis), which spuriously correlates with the \emph{contradiction} label.
We use the standard train / valid / test splits from \citet{sagawa2020group}.

\subsection{CivilComments-WILDS}\label{sec:cc_breakdown}

We use the CivilComments-WILDS dataset from \citet{koh2021wilds}, which is derived from the Jigsaw dataset \citep{borkan2019nuanced}.
Given a real online comment, the task is to predict whether the comment is \emph{toxic} or \emph{not toxic}.
The spurious attribute $\spuattribute$ is an 8-dimensional binary vector, where each entry is a binary indicator of whether the following 8 demographic identities are mentioned in the online comment:
\emph{male}, \emph{female}, \emph{LGBTQ}, \emph{Christian}, \emph{Muslim}, \emph{other religion}, \emph{Black}, and \emph{White}.

Following \citet{koh2021wilds}, we consider the 16 \emph{potentially overlapping} groups equal to (\emph{identity}, \emph{toxic}) and (\emph{identity}, \emph{not toxic}) for all 8 identities.
We use the standard train / valid / test splits from \citet{koh2021wilds}.

\section{Additional Experimental Results}\label{sec:extra-experiments}

\begin{table}[]
\centering
\resizebox{\columnwidth}{!}{\begin{tabular}{ccc}
\toprule
Group & Enrichment & ERM test acc. \\
  \cmidrule(lr){1-1} \cmidrule(lr){2-2} \cmidrule(lr){3-3}
(muslim, toxic) & 8.58x & 62.1\% \\
(christian, toxic) & 8.58x & 57.4\% \\
(LGBTQ, toxic) & 8.58x & 72.1\% \\
(other religion, toxic) & 8.56x & 62.9\% \\
(black, toxic) & 8.53x & 74.6\%\\
(white, toxic) & 8.49x & 68.6\% \\
(female, toxic) & 8.49x & 64.7\% \\
(male, toxic) & 8.48x & 66.6\% \\
(white, non-toxic) & 0.09x & 84.7\% \\
(LGBTQ, non-toxic) & 0.07x & 82.2\% \\
(black, non-toxic) & 0.07x & 74.6\% \\
(male, non-toxic) & 0.06x & 93.0\% \\
(muslim, non-toxic) & 0.05x & 89.4\% \\
(female, non-toxic) & 0.05x & 94.7\% \\
(other religion, non-toxic) & 0.04x & 93.4\% \\
(christian, non-toxic) & 0.03x & 96.3\% \\
\bottomrule
\end{tabular}
}
\center\small
\vspace{-0.3cm}
    \caption{CivilComments error set breakdowns.
    }
\label{tab:civilcomments-breakdown}
\end{table}

\subsection{Error set analysis for CivilComments}
We include the CivilComments error set analysis in \reftab{civilcomments-breakdown} for space constraints.

\subsection{Additional analysis}
Below, we present a series of analyses that involve partitioning the dataset into two groups: groups in which spurious correlation holds with $y=a$, and groups in which spurious correlation does not h  old with $9\neq a$.
For this investigation, we focus on Waterbirds and CelebA, where all groups can be clearly partitioned as above because we consider binary classification tasks with binary spurious attributes.
In Waterbirds, the $y=a$ groups are waterbirds on water backgrounds and landbirds on land backgrounds; the $y\neq a$ groups are waterbirds on land backgrounds and landbirds on water backgrounds.
In CelebA, the $y=a$ groups are blond females and non-blond males; the $y\neq a$ groups are non-blond females and blond males.
In contrast, it is unclear how to partition the groups as above in MultiNLI, in which we consider a multi-class classification problem, and in CivilComments-WILDS, in which we have multiple spurious a  ttributes corresponding to different demographic identities.

\paragraph{Impact of $y=a$ and $y\neq$ examples in the error set.}
\begin{figure}
  \centering
  \includegraphics[width=0.9\columnwidth]{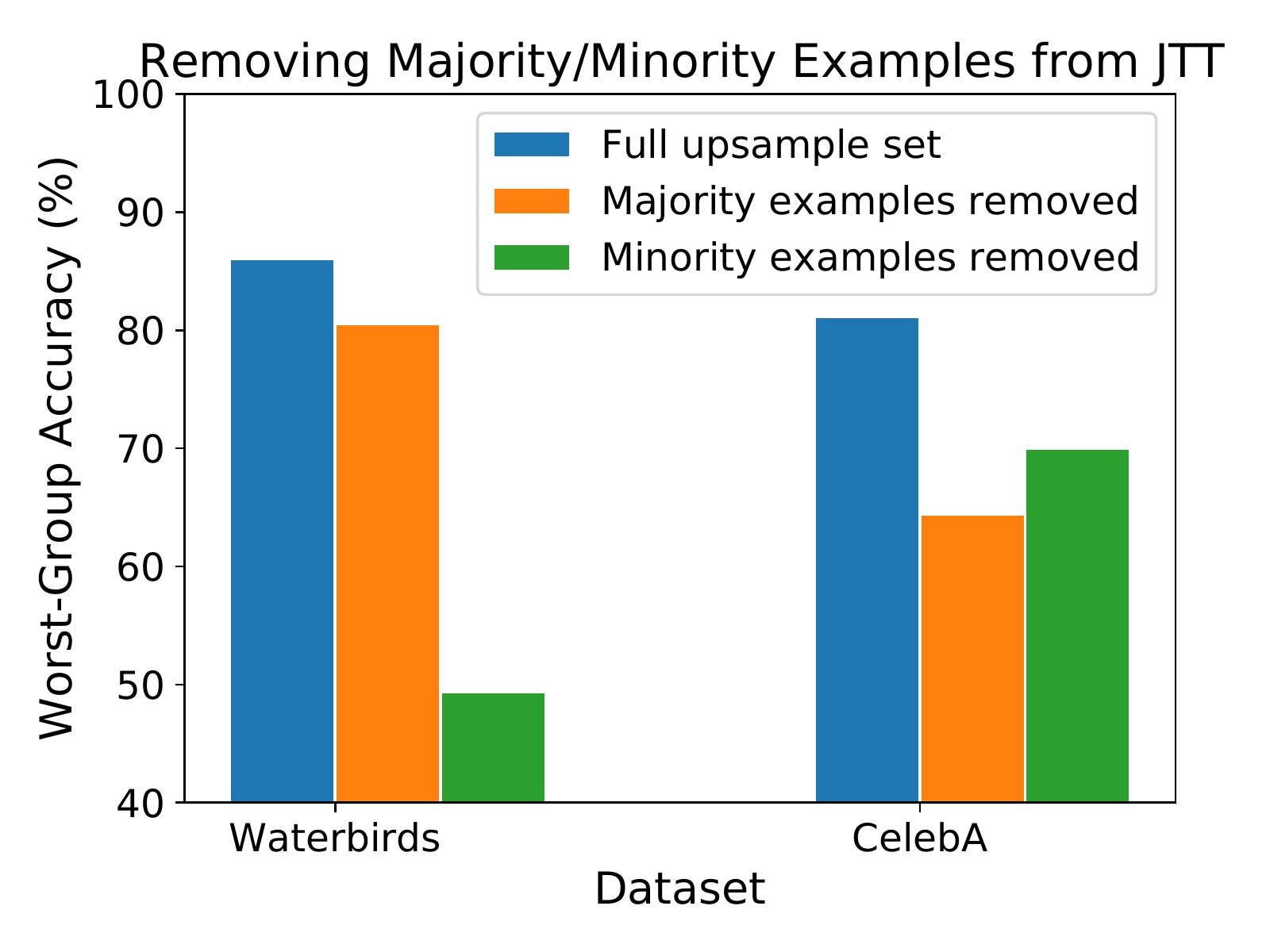}
  \vspace{-0.2cm}
  \caption{
      Effect on worst-group accuracy of removing the $y=a$ and $y\neq a$ examples from \ours's error set.
      Both upsampling $y=a$ examples and upsampling $y\neq a$ examples substantially contribute to improving worst-group accuracy.
  }
 \label{fig:majority_minority}
\end{figure}

We first study how worst-group accuracy changes when we remove $y=a$ examples or $y\neq a$ examples from the error set, as summarized in \reffig{majority_minority}.
In both datasets, removing either the $y=a$ or $y\neq a$ examples from the error set significantly decreases worst-group accuracy.
While this reduction in worst-group accuracy could stem from the fact that we consider a fixed set of hyperparameters including the upweight factor (which was tuned for \ours with the full error set),   it is possible that both $y=a$ and $y\neq a$ examples contribute to the improvement in worst-group accuracy.
In particular, because both datasets have substantial label imbalance, it is expected that upweighting groups from rare labels is important to perform well on all groups, and in fact, groups with $y\neq a$ and with rare $y$ are upweighted as discussed in \refsec{error_set}.

\begin{figure}
  \centering
  \includegraphics[width=0.9\columnwidth]{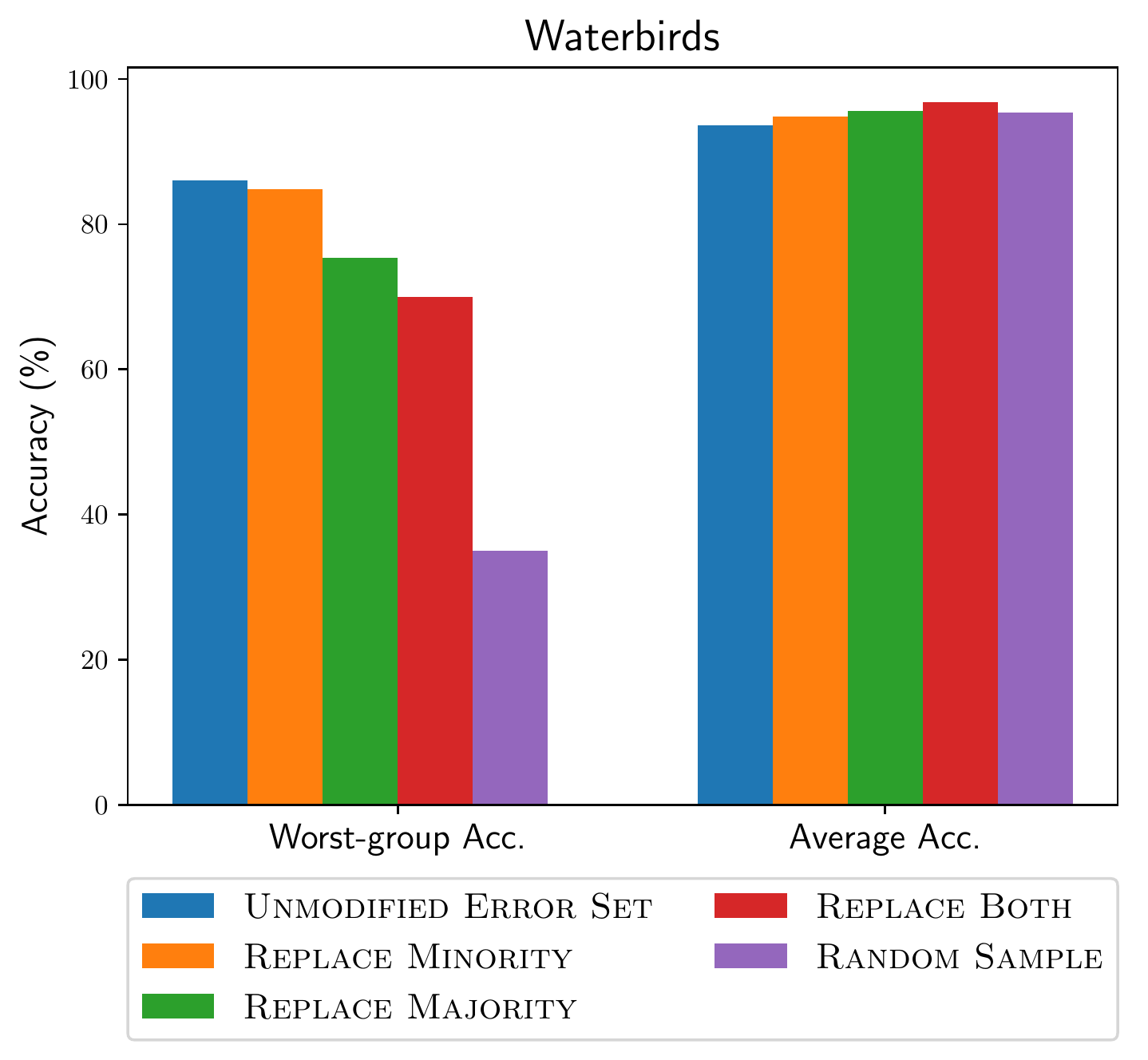}
  \vspace{-0.2cm}
  \caption{
      Effect of replacing the $y\neq a$ and $y=a$ examples in \ours's error set with randomly-selected $y\neq a$ and $y=a$ examples on Waterbirds.
     Worst-group accuracy decreases when replacing either the $y=a$ or $y\neq a$ examples, suggesting that \ours successfully automatically identifies informative examples that improve worst-group accuracy when upsampled.
  }
 \label{fig:error_set_ablation}
\end{figure}

Next, we explore if the particular $y=a$ or $y\neq a$ examples that \ours upsamples is important, or if upsampling \emph{any} collection of examples in these groups yields high worst-group accuracy.
To do this, we study how average and worst-group accuracies change when we replace examples in \ours's error set with randomly selected examples from specific groups.
Specifically, we study what happens when we upsample the following four variants of \ours's error set:

\begin{itemize}
  \item \textsc{Replace $y\neq a$}: We replace the $y\neq a$ examples in the error set with an equal number of randomly selected $y\neq a$ examples, leaving the $y=a$ examples in the error set uncha  nged.
  \item \textsc{Replace $y=a$}: We replace the $y=a$ examples in the error set with an equal number of randomly selected $y=a$ examples, leaving the $y\neq a$ examples in the error set unchanged.
  \item \textsc{Replace both}: We replace both the $y\neq a$ examples in the error set with an equal number of randomly selected $y\neq a$ examples, and the $y=a$ examples in the error set with an e  qual number of randomly selected $y=a$ examples.
  \item \textsc{Random sample}: We replace all examples in the error set with an equal number of randomly selected examples.
  This yields a different fraction of $y\neq a$ examples in the error set compared to \textsc{Replace both}.
\end{itemize}

\reffig{error_set_ablation} compares upsampling these variants of the error set with the original unmodified error set (\textsc{unmodified error set}).
Compared to upsampling the original error set, upsampling \textsc{Replace $y\neq a$} slightly decreases worst-group accuracy and leaves average accuracy unchanged.
This suggests that upsampling most $y\neq a$ examples helps improve worst-group accuracy, though the particular $y\neq a$ examples \ours identifies in the error set are still slightly better than rand  om.
On the other hand, upsampling \textsc{Replace $y=a$} significantly decreases worst-group accuracy, although it slightly improves average accuracy compared to the original error set.
This suggests that the particular $y=a$ examples \ours identifies in the error set are important for improving worst-group accuracy. This could be because the label balance within upsampled $y=a$ changes, or for other reasons.
Finally, the low worst-group and average accuracies of both \textsc{Replace both} and \textsc{Random sample} show that merely upsampling random $y=a$ and $y\neq a$ examples is insufficient to achieve   high worst-group accuracy.

\paragraph{Upsampling $y\neq a$ groups.}
We present the performance of a simple baseline, in which we upweight $y \neq a $ examples using ground-truth group annotations, in \reftab{upsample_minority}.
While \spuoracle improves worst-group error over ERM, this baseline is limited in a few ways.
First, $y\neq a$ groups are not necessarily groups with the worst accuracies or smallest number of examples, for example due to label imbalance.
So while we $y\neq a$ examples are counter-examples to the spurious correlations, it's not necessarily expected that they improve the worst-group performance well.
Secondly, in the presence of ground-truth examples, it is possible to reweight each of the groups independently, rather than reweighting $y=a$ and $y\neq a$ groups.
Prior work has observed much higher worst-group performance by reweighting the groups than \spuoracle \citep{sagawa2020group}.

\begin{table}
\centering
\footnotesize
\resizebox{\columnwidth}{!}{\begin{tabular}{ccccc}
\toprule
& \multicolumn{2}{c}{Waterbirds} & \multicolumn{2}{c}{CelebA} \\
& Avg. & Worst-group & Avg. & Worst-group \\
\cmidrule(lr){2-3} \cmidrule(lr){4-5}
\spuoracle & 96.7\% & 75.9\% & 93.4\% & 57.2\% \\
\ours & 93.3\% & \textbf{86.7\%} & 88.0\% & \textbf{81.1\%} \\
\bottomrule
\end{tabular}
}
\center
\caption{
Average and worst-group test accuracies. \spuoracle, which upsamples $y\neq a$ examples, achieves higher worst-group accuracy than ERM, but lower than \ours.
}
\label{tab:upsample_minority}
\end{table}

\end{document}